\documentclass[runningheads]{llncs}
\usepackage{beamerarticle}
\usepackage[utf8]{inputenc}

\usepackage{amsmath,amsthm}
\usepackage{graphicx}
\usepackage{amssymb}
\usepackage{bm}
\usepackage[shortlabels]{enumitem}
\usepackage{graphicx}
\usepackage{csquotes}
\usepackage{float}
\usepackage{booktabs} 
\usepackage{csquotes}
\usepackage{eurosym}
\usepackage{adjustbox}
\usepackage{abstract}
\usepackage{colortbl}

\usepackage{graphicx}
\usepackage{capt-of}
\usepackage{caption}
\usepackage[breaklinks=true]{hyperref}

\usepackage[sorting=none]{biblatex}
\begin{document}

\theoremstyle{definition}
\newtheorem*{exmp}{Example}

\title{Logic Constraints to Feature Importances}
\author{
Nicola Picchiotti\inst{1, 2}\orcidID{0000-0003-3454-7250} 
\and
Marco Gori\inst{1, 3}
}

\authorrunning{N. Picchiotti et al.}
%
\institute{SAILAB, University of Siena 
\url{http://sailab.diism.unisi.it} \and
 University of Pavia \email{nicola.picchiotti01@universitadipavia.it}
 \and
 MAASAI, Universit\`e C\^ote d'Azur
\email{marco@diism.unisi.it}
}
\maketitle

\begin{abstract}
In recent years, Artificial Intelligence (AI) algorithms have been proven to outperform traditional statistical methods in terms of predictivity, especially when a large amount of data was available. Nevertheless, the "black box" nature of AI models is often a limit for a reliable application in high-stakes fields like diagnostic techniques, autonomous guide, etc. Recent works have shown that an adequate level of interpretability could enforce the more general concept of model trustworthiness \cite{doshi2017towards}. The basic idea of this paper is to exploit the human prior knowledge of the features' importance for a specific task, in order to coherently aid the phase of the model's fitting. This sort of "weighted" AI is obtained by extending the empirical loss with a regularization term encouraging the importance of the features to follow predetermined constraints. This procedure relies on local methods for the feature importance computation, e.g. LRP, LIME, etc. that are the link between the model weights to be optimized and the user-defined constraints on feature importance. In the fairness area, promising experimental results have been obtained for the Adult dataset. Many other possible applications of this model agnostic theoretical framework are described.
\end{abstract}

\section{Introduction}

Trustworthiness of an Artificial Intelligence (AI) model, i.e. the stability of the performances under many possible future scenarios, is a fundamental requirement for the real world applications. The topic has became particularly relevant in the last decades, since technology development and data availability led the adoption of models more and more complex, widen the gap between performances on train/test data and reliability of the models.

Model trustworthiness is usually linked to other factors, including the interpretability of the algorithm, the stationary of data, the possible bias in the data, etc. (\cite{lipton2018mythos}, \cite{ribeiro2016should}). Especially in the field of interpretability, many work has been done in order to explain and interpret the models developed by AI in a human comprehensible manner. The main reason behind these effort is that the human experience and its capacity for abstraction allow to monitor the process of the model decisions in a sound way, trying to mitigate the risk of data-driven models.

Anyway, an effective interaction between the model and the human is still lacking and mostly of the current machine learning approaches tends to rely too heavily on training/testing data. On the other hand, sources of knowledge like domain knowledge, expert opinions, understanding from related problem etc. could be very important for a better definition of the model.

Here we present a novel framework trying to bridge the gap between data-driven optimization and human high-level domain knowledge. The approach provides for the inclusion of the human understanding of the relevance/importance of the input features. The basic idea is to extend the empirical loss with a regularization term depending on the constraints defined by the apriori knowledge on the importance of the features. We provide experimental results on the fairness topic.

\section{Bibliographic review}

There are few existing feature weighting approaches aimed at improving the performances of machine learning models. In \cite{al2011empirical} the author exploits weak domain knowledge in the form of feature importance to help the learning of Importance-Aided Neural Networks (IANN). The feature importance is based on the absolute weight of the first hidden layer neurons of the network. IANN is successfully applied in \cite{diersen2011classification}. In \cite{zhang2010ontology} an ontology-based clustering algorithm is introduced along with a feature weights mechanism able to reflect the different features' importance. \cite{peng2018novel} uses both correlation and mutual information to weight the features for the algorithms SVM, KNN, and Naive Bayes. Instead, in order to accelerate the learning process, in \cite{iqbal2011using} the algorithm is required to match the correlation between the features and the predictive function with the empirical correlation.

Anyway, none of the previous works define a general framework including the knowledge on the importance of the input features in the framework of explainable machine learning.

\section{Mathematical setting of feature importance}
In this section, we review the existing approaches aimed at assigning an importance score for each feature of a given input example in relation to the task of the model, i.e. the so-called local explainable methodologies. It is worth mentioning that the importance of a feature is one of the most used strategies to gain local explainability from an opaque machine learning model.\\\\
Let us consider a \textit{predictor function} $\tilde{f}$ going from the d-dim feature space $\mathcal{X} = \mathcal{X}_1 \times \mathcal{X}_2 ... \times \mathcal{X}_d$ to the 1-dim target space $\mathcal{Y}$:
$$
\tilde{f}: \mathcal{X} \rightarrow \mathcal{Y}.
$$
Such a predictor function is the output of a learner:
$$
\mathcal{L} :
\left(
\mathcal{X}^{n} \times \mathcal{Y}^n
\right)
\rightarrow
\left(
\mathcal{X} \rightarrow \mathcal{Y}
\right),
$$

able to process a supervised dataset $\mathcal{D}=\{X, Y\}$ where $X \in \mathcal{X}^{n}$ and $Y \in \mathcal{Y}^{n}$ with $n$ instances.
In order to fix the idea, we can think of $\mathcal{L}$ as a Deep Neural Network providing the predictor function $\tilde{f}$.

\begin{definition}[Local feature importance]
The local feature importance is a function mapping a predictor $\tilde{f}$ and a single instance $\boldsymbol{x} \in \mathcal{X}$ to a $d$-dim vector of real values in the range $[0,1]$:
$$
\boldsymbol{I}(\tilde{f}, \boldsymbol{x}): 
\left(
\mathcal{X} \rightarrow \mathcal{Y}
\right)
\times
\mathcal{X}
\rightarrow
[0, 1]^d
$$
\newcommand{\at}[2][]{#1|_{#2}}
\end{definition}
The local feature importance is a measure of how much the model relies on each feature for the prediction $\tilde{f}(\boldsymbol{x})$ made by $\tilde{f}$ on the particular pattern $\boldsymbol{x}$. Basically, the quantity $I_i(\tilde{f}, \boldsymbol{x})$ tells us how much the $i$-th feature contributes with respect to the others for a specific prediction. In the limit cases of $I_i(\boldsymbol{x})=0$ or $I_i(\boldsymbol{x})=1$ the feature $i$ can be considered respectively useless or the most important one for the prediction done by the predictor on the pattern $\boldsymbol{x}$.

\begin{exmp}
Given a linear predictor $\tilde{f}(\boldsymbol{x}) = \sum_{i=1}^d w_i x_i$, the function 
$$
\boldsymbol{I}(\tilde{f}, \boldsymbol{x})= \frac{| \boldsymbol{w} |}{max_{i \in [1, d]}|w_i|}
$$ 
is a local feature importance function.
\end{exmp}

\begin{definition}[Local feature importance methods]
Local feature importance methods are methods that given a predictor $\tilde{f}$, with its learner and the dataset, computes a local feature importance function $\boldsymbol{I}(\tilde{f}, \boldsymbol{x})$.
\end{definition}

The existing methodologies for the computation of feature importance are reviewed in \cite{guidotti2018survey} and \cite{arrieta2020explainable}. \textit{Permutation feature importance} methods quantify the feature importance through the variation of a loss metric by perturbing the values of a selected feature on a set of instances in the training or validation set. The approach is firstly introduced in \cite{breiman2001random} for random forest and in \cite{recknagel1997artificial} for Neural Network. Other methods, such as \textit{class model visualization} \cite{simonyan2013deep}, compute the partial derivative of the score function with respect to the input, and \cite{montavon2018methods} introduce expert distribution for the input giving \textit{activation maximization}. In the paper \cite{shrikumar2017learning} the author introduces \textit{deep lift} and computes the discrete gradients with respect to a baseline point, by backpropagating the scoring difference through each unit. Instead, integrated gradients \cite{sundararajan2016gradients} cumulates the gradients with respect to inputs along the path from a given baseline to the instance.

A set of well known methods called \textit{Additive Feature Attribution Methods} (AFAM) defined in \cite{lundberg2017unified} rely on the redistribution of the predicted value $\hat{f}(\boldsymbol{x})$ over the input features. They are designed to mimic the behaviour of a predictive function $\hat{f}$ with a surrogate Boolean linear function $g$. This surrogate function takes values in a space of the transformed vector of the input features: $\boldsymbol{x}^{\prime} =h(\boldsymbol{x}) \in [0, 1]^d$: 
$$
\hat{f}(\boldsymbol{x}) \approx g(\boldsymbol{x}^{\prime}) = \phi_0 + \sum_{i=1}^d \phi_i x_i^{\prime}.
$$
Keeping the notation introduced above, by defining
$$
\boldsymbol{I}(\hat{f}, \boldsymbol{x}) = \frac{| \boldsymbol{\phi} |}{max_{i \in [1, d]}|\phi_i|},
$$
we have a local feature importance method.
\newcommand{\at}[2][]{#1|_{#2}}

Among the additive feature attribution methods, the popular \textit{LIME} (Local Interpretable Model-agnostic Explanations) \cite{ribeiro2016should} builds the linear approximated model with a sampling procedure in the neighborhood of the specific point. By considering proper weights to the linear coefficients of LIME, the author in \cite{lundberg2017unified} demostrated that \textit{SHAP} (SHapley Additive exPlanation) is the unique solution of additive feature attribution methods granting a set of desirable properties (local accuracy, missingness and consistency). This last method lays the foundation on the \textit{Shapley} method introduced in \cite{strumbelj2010efficient} and \cite{vstrumbelj2014explaining} for solving the problem of redistributing a reward (prediction) to a set of player (features) in coalitional game theory framework.
Finally, \textit{Layer-wise Relevance Propagation} (LRP) in \cite{bach2015pixel} backpropagates the prediction along the network, by fixing a redistribution rule based on the weights among the neurons.\\\\
The set of feature importance methods, along with the data type and the model to which the method is referred are reported in Table \ref{fim}. For a tabular dataset, the feature importances are usually represented as a rank reported in a histogram. For images or texts, the subset of the input which is mostly in charge of the predictions gives rise to saliency masks; for example they can be parts of the image or a sentences of a text.

\begin{table*}[h!]
    \caption{Review of the most known feature importance methods in Explainability. TAB: Tabular dataset, IMG: Images, AGN: Model agnostic methodology, NN: Neural Network, TE: Tree Ensemble, AFAM: Additive Feature Attribution Methods.}
  \centering
    \begin{tabular}{ccccc}
    \toprule
    \textbf{Method} & \textbf{Data Type} & \textbf{Model} & \textbf{Reference} & \textbf{AFAM} \\
    \midrule
    \rowcolor[rgb]{ .851,  .851,  .851} SHAP  & ANY   & AGN   & \cite{lundberg2017unified} & v \\
    LIME  & ANY   & AGN   & \cite{ribeiro2016should} & v \\
    \rowcolor[rgb]{ .851,  .851,  .851} Shapley value & TAB   & AGN   & \cite{strumbelj2010efficient} \cite{vstrumbelj2014explaining} & v \\
    Permutation feature importance & ANY   & NN/TE & \cite{breiman2001random} \cite{recknagel1997artificial} & - \\
    \rowcolor[rgb]{ .851,  .851,  .851} class model visualization & IMG   & NN    & \cite{simonyan2013deep} & - \\
    activation maximization & IMG   & NN    & \cite{montavon2018methods} & - \\
    \rowcolor[rgb]{ .851,  .851,  .851} LRP   & ANY   & NN    & \cite{bach2015pixel} & v \\
    Taylor Decomposition & ANY   & NN    & \cite{bach2015pixel} & - \\
    \rowcolor[rgb]{ .851,  .851,  .851} DeepLift & ANY   & NN    & \cite{shrikumar2017learning} & v \\
    Integrated Gradients & ANY   & NN    & \cite{sundararajan2016gradients} & - \\
    \rowcolor[rgb]{ .851,  .851,  .851} GAM   & ANY   & AGN   & \cite{lou2012intelligible} \cite{lou2013accurate} & - \\
    \bottomrule
    \end{tabular}%

  \label{fim}%
\end{table*}%


\section{Constraints to feature importance}
The overall goal of the present work is to define a framework where the local importance of the model's features can be constrained to specific intervals. We introduce a novel regularization loss term $L_I$, related to the not fulfillment of the feature importance's constraints:
\begin{equation}\label{loss}
L_r(\boldsymbol{x}, \boldsymbol{w}) + L_I(\boldsymbol{I}(\tilde{f}(\boldsymbol{w}, \cdot), \boldsymbol{x})),
\end{equation}
where $L_r$ is the usual empirical risk loss and $\boldsymbol{I}$ the $d$-dim vector of importances. It is worth observing that in Eq.~\eqref{loss} we explicated the dependence of the importance on the structure of the black box model via the weights of the model $\boldsymbol{w}$. \\\\
Let us suppose a First Order Logic (FOL) formula $E(\boldsymbol{I}(\tilde{f}, \boldsymbol{x}))$ with variable $\boldsymbol{x} = [x_1, x_2 ..., x_d]$ containing an apriori statement with inequalities on the features' importances. For example, we could require that, for every $\boldsymbol{x}$, both the feature $1$ and the feature $2$ should not be important for the prediction function to properly work:
\begin{equation}\label{folformula}
\forall \boldsymbol{x}:\,\,\, I_1(\tilde{f}, \boldsymbol{x})< c_1 \land I_2(\tilde{f}, \boldsymbol{x}) < c_2,
\end{equation}

with $c_1 \in [0, 1]$ and $c_2 \in [0, 1]$.

In order to treat the logic formula with real value functions, each inequality of the FOL formula can be transformed into a new variable $l_{i, c_i} \in [0, 1]$ through the following transformation:

\begin{equation}\label{ineqtransform}
\begin{aligned}
I_i(\tilde{f}, \boldsymbol{x})<c_i \, \, \longrightarrow \, \, l_{i, c_i}(\boldsymbol{x}) = \frac{\max(I_i(\tilde{f}, \boldsymbol{x}) - c_i,\, 0)}{1-c_i}.
\end{aligned}
\end{equation}

\begin{figure}[!htb]
    \begin{center}
        \includegraphics[width=7cm]{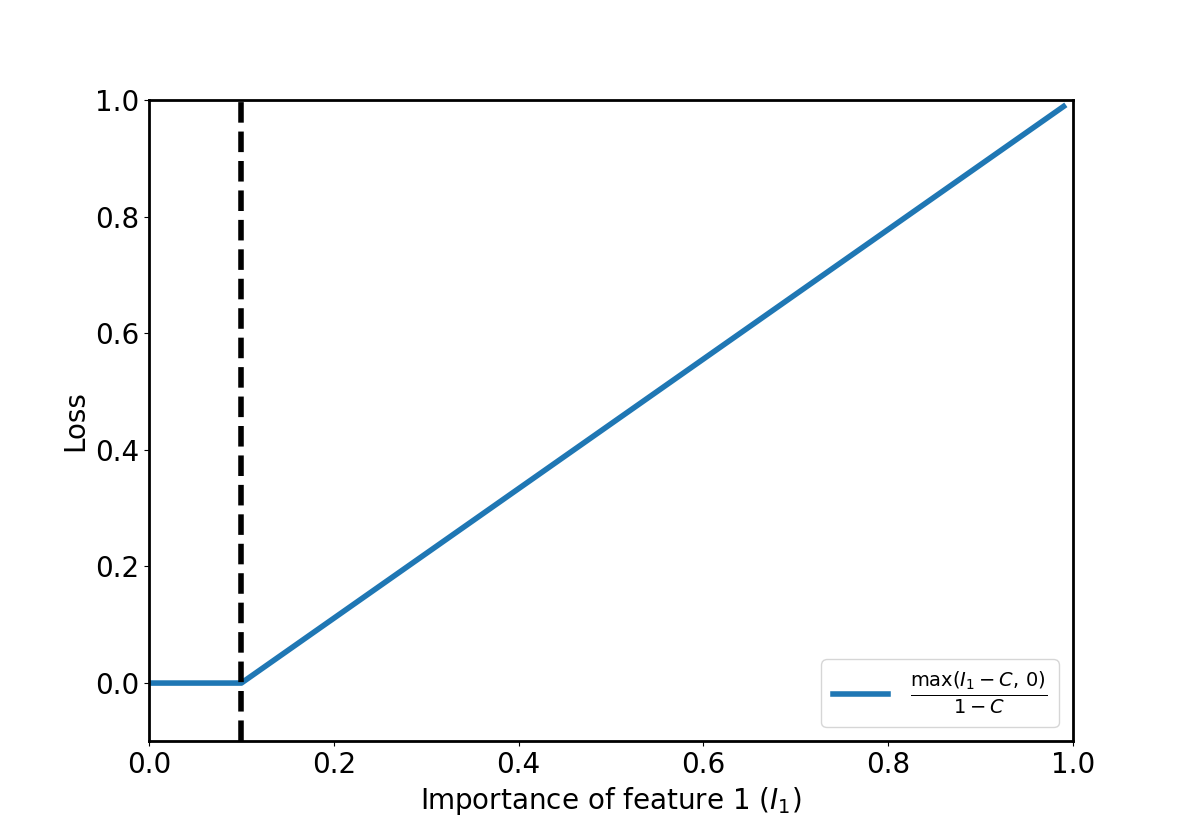}
    \end{center}
    \caption{Example of loss as for the inequality on the importance of the constrained feature for $I_i < 0.1$.}
    \label{figureloss}
\end{figure}

Although Eq.~\eqref{ineqtransform} is a quite natural choice for an increasing function from $0$ to $1$, other choices are possible. In Figure~\ref{figureloss} we represent the variable $l_{i, c_i}(\boldsymbol{x})$ of Eq.~\eqref{ineqtransform} for the case $c_i=0.1$ of a generic feature.

So, thanks to Eq.~\eqref{ineqtransform}, the aforementioned FOL formula Eq.~\eqref{folformula} can be written as:

$$
\forall \boldsymbol{x}:\,\,\, l_{1, c_1}(\tilde{f}, \boldsymbol{x}) \land l_{2, c_2}(\tilde{f}, \boldsymbol{x}).
$$

Then, we exploit the framework of t-norm fuzzy logic that generalizes Boolean logic to variables assuming values in $[0, 1]$. We can convert the formula depending on the losses $E(\boldsymbol{l}(\tilde{f}, \boldsymbol{x}))$ by exploiting a T-norm $t$ in the following:
$$
\Phi_{\forall}(\boldsymbol{l}(\tilde{f}, \mathcal{X})) = \frac{1}{|\mathcal{X}|} \sum_{\boldsymbol{x} \in \mathcal{X} } t_{E} \left( \boldsymbol{l}\left( \tilde{f}, \boldsymbol{x} \right) \right),
$$
that is an average over the t-norm of the truth degree when grounded $\boldsymbol{x}$ over its domain. Then, a loss term can be defined by exploiting the logic constraints, e.g.
$$
L_I(\boldsymbol{I}(\tilde{f}, \mathcal{X})) = \lambda (1 - \Phi_{\forall}(\boldsymbol{I}(\tilde{f}, \mathcal{X})) ),
$$
where $\lambda$ is the strength of the regularization.

Finally, the partial derivative of the logic part of the loss $L_I$ with respect to the $j$-th weight of the $i$-th importance loss function is
$$
\frac{\partial L_I}{\partial w_{i, j}} = 
\sum_k
\frac{\partial L_I}{\partial \Phi_k} 
\frac{\partial \Phi_k}{\partial l_i}
\frac{\partial l_i}{\partial w_{i, j}},
$$
and the derivative to being evaluated is:
$\frac{\partial l_i}{\partial w_{i, j}}$.
By resuming, the scheme is the following:
\begin{enumerate}
    \item write the FOL formula depending on the feature importance, in turn, depending on the model weights through the chosen feature importance method;
    \item convert the inequality terms $I_i<c_i$ into loss terms $l_{i, c_i}$;
    \item convert the FOL formula with the t-norm into an overall loss term;
    \item the loss term is optimized in an iterative process by computing the importance at each step of the algorithm.
\end{enumerate}

\section{Fairness through feature importance constraints}

Fairness is a natural field where the constraints to feature importance can be applied. In the following, we resume the principal fairness measures and we discuss how they can be translated by using our proposed scheme based on Constraints to Feature Importance, denoted hereafter as CTFI.

The \textit{Demographic Parity} (DP) fairness metric is satisfied when, given the random variable $\tilde{Y}$ representing the binary predictor $\tilde{f}$ and a protected Boolean feature $x_s$ we have: 
$$
P(\tilde{Y}=1 | x_s = 0) = P(\tilde{Y}=1| x_s = 1).
$$
DP is a very strong requirement: groups based on a sensitive feature, e.g. black and white, should have the same rate of positive prediction, even if differences are present.

DP can be translated into a constraint, where the importance of the protected feature $s$ needs to be lower than a given threshold $c \in [0,1]$:
\begin{equation}\label{cofei_dp}
\forall \boldsymbol{x} \;\;   I_s(\boldsymbol{x})<c.
\end{equation}

The possible well-known issue of \textit{unfairness due to correlated features} (see for instance \cite{calders2013unbiased}) can be potentially solved by setting a constraint also for the features that are correlated with the protected one. Obviously, the regularization strength $\lambda_i$ of the $i$-th correlated feature should be lower with respect to $\lambda_s$, for instance given by the product of $\lambda_i$ and the Pearson correlation between the $s$-th and $i$-th feature, i.e., $\rho_{s, i}$:
\begin{equation}\label{corr}
\lambda_i = \lambda_s \cdot \rho_{s, i}.
\end{equation}
The advantage of this formulation is that the constraints are smooth between $0$ and $1$, and can be used both with binary and continuous features. 

A measure of discrepancy from DP, which will be useful for the experimental part, is the \textit{Disparate impact} (DI):

\begin{equation}\label{di}
\text{DI} = \frac{P(\tilde{Y} = 1|X_s = 0)}{P(\tilde{Y} = 1|X_s = 1)}.
\end{equation}



A possible relaxation of DP is where we grant that the protected attribute $x_s$ can be used to discriminate among groups that are actually different in the ground truth label $y$, i.e.,  between $y=0$ and $y=1$, but not within each one. This is called \textit{Equalized odd} (EOD) and is described in paper \cite{hardt2016equality}. We say that a predictor $\tilde{Y}$ satisfies EOD if $\tilde{Y}$ and $X_s$ are independent, conditional on $Y$:
$$
P( \tilde{Y} = 1 | X_s = 0; Y=1) = P( \tilde{Y} = 1 | X_s = 1; Y=1),
$$
$$
P( \tilde{Y} = 1 | X_s = 0; Y=0) = P( \tilde{Y} = 1 | X_s = 1; Y=0).
$$

EOD, too, can be easily written in the framework of CTFI as:
$$
\forall \; \boldsymbol{x} \in \mathcal{X} \, | \, y(\boldsymbol{x})=1 \;   : I_s(\boldsymbol{x})<c,
$$
$$
\forall \; \boldsymbol{x} \in \mathcal{X} \, | \,  y(\boldsymbol{x})=0 \;   : I_s(\boldsymbol{x})<c.
$$
with $c \in [0, 1]$.

A quite natural measure of discrepancy from EOD is the \textit{average equality of odds difference} (EO):

\begin{align}\label{eo}
\begin{split}
\text{EO} &=\frac{ P( \tilde{Y} = 1 | X_s = 0; Y=1) - P( \tilde{Y} = 1 | X_s = 1; Y=1)}{2}\\
& + \frac{P( \tilde{Y} = 1 | X_s = 0; Y=0) - P( \tilde{Y} = 1 | X_s = 1; Y=0)
}{2} .
\end{split}
\end{align}

Finally, another measure of fairness discrepancy defined in \cite{kusner2017counterfactual} is \textit{counterfactual fairness difference} (CF):

\begin{equation}\label{cf}
\text{CF} = P( \tilde{Y}_{x_s \leftarrow 0} = 1 | X_s = 1) - P(\tilde{Y} = 1 | X_s = 1),
\end{equation}

where the idea is to evaluate the differences of the prediction's probabilities by changing the protected feature of the patterns from $1$ to $0$.

\section{Toy example: constraint of the form \texorpdfstring{$I_i(\boldsymbol{x})<c$}{Lg}}

As a toy example useful to test the effectiveness of the proposed scheme, we used the German credit risk dataset ($1000$ instances), available in \cite{Dua:2019}, containing information about bank account holders and a binary target variable denoting the credit risk. The considered features are reported in Table~\ref{german}.

\begin{table}[!htb]
  \caption{Features for the German credit risk dataset.}
  \label{german}
  \centering
  \begin{tabular}{ccc}
    \toprule
    Feature     & Description     & Range \\
    \midrule
    Age & Age of the costumer & numerical [19, 74]     \\
    Job     &  Job qualification & ordinal [0, 3]      \\
    Amount     &    Credit Amount (\EUR{}) of the loan     & numerical  \\
    Duration     & Duration (year) of the loan &  numerical [4, 72] \\
    Gender     & Male (1) Vs Female (0)  & Boolean
  \end{tabular}
\end{table}

We exploited a neural network with one hidden layer and $16$ neurons. The learning rate of SGD is $0.01$ with $10$ epochs. The activation function is ReLU and the loss is given by the binary cross-entropy.

\begin{figure}[!htb]
    \begin{center}
        \includegraphics[width=10cm]{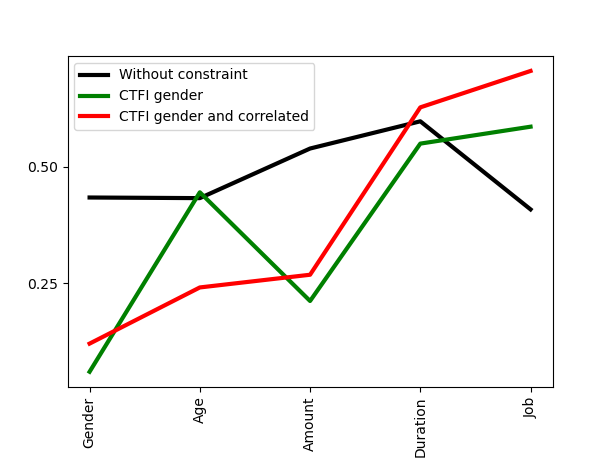}
    \end{center}
    \caption{Feature importance (LRP) for the original model (black line), for the model with the constraints on the \textit{gender} feature (green line) and that constraining also the correlated features (red line).}
    \label{fi}
\end{figure}

After the training phase, the Layer-wise Relevance Propagation method has been applied to the instances of the testing set (50\% of the overall samples) for computing the feature importances. The black line in Figure~\ref{fi} reports the average feature importance computed with LRP. We observe that the most relevant feature is the \textit{duration} of the loan, followed by the \textit{amount} and the \textit{gender}. 

Let us introduce a constraint to the importance of \textit{gender} feature that we want to be less-equal than zero (see Eq.~\eqref{cofei_dp}), with a regularization $\lambda=0.05$. As expected, we observe (green line in Figure~\ref{fi}) that the \textit{gender} feature has become useless for the model predictions. Basically, the model found another solution, by giving more importance to other features, e.g. the \textit{job}.

Furthermore, we computed the correlation matrix between the different features and we found out that, for instance, the \textit{age} feature is correlated with the \textit{gender} (with a Pearson correlation coefficient of $\rho_{\text{gender}, \text{age}}=16\%$). So, in the third experiment we constrained also the other features, by using different regularization strength given by: 
$$\lambda_i =\lambda \cdot \rho_{\text{gender}, \,i}$$
for the $i$-th feature (see Eq.~\eqref{corr}). We note from the results reported in Figure~\ref{fi} with a red line, that the correlated \textit{age} feature decreases its importance, coherently with the expectation.

\section{Fairness through constraints to feature importance}

In this section we report the results of the experimental part related to fairness. We tested the CTFI scheme proposed in the previous section to the \textit{Adult income} data set, also considered by \cite{kamishima2011fairness}. It contains $48,842$ instances with $12$ attributes (see Table~\ref{adult} for the description) and a binary classification task for people earning more or less than $\$50,000$ per year. The protected attribute we will examine is the \textit{race}, categorized as white and non-white. In order to better evaluate the fairness metrics with a uniform test set, the dataset has been balanced and the chosen split of training/test is 50\%. The model is a Neural Network with one hidden layer and $4$ neurons. The learning rate is $0.1$, the number of epochs is $10$ and the batch size is fixed to $1$ in order to compute the local feature importance for each analyzed pattern. The activation function is the ReLU function and the loss is given by the binary cross-entropy.

\begin{table}[!htb]
  \caption{Features for the Adult income dataset.}
  \label{adult}
  \centering
  \begin{tabular}{ccc}
    \toprule
    Feature  & Range \\
    \midrule
    Age & numerical [19, 74]     \\
    Race & Boolean: white Vs non-white \\
    Sex & Boolean: female Vs male \\
    Education &  ordinal: [1, 5] \\ 
    Native-country &  Boolean: US Vs other \\
    Marital-status &  Boolean: single Vs couple \\
    Relationship &  ordinal: [1, 5] \\
    Employment type &  ordinal: [1, 5] \\
    fnlwgt & continuous\\
    Capital loss & Boolean: Yes Vs NO \\
    Capital gain & Boolean: Yes Vs NO \\
    hours-per-week & continuous\\
  \end{tabular}
\end{table}

We used the constraint defined in Eq.~\eqref{cofei_dp} with $c=0$ for the \textit{race} feature; whereas as a fairness metric we consider both the disparate impact (DI) defined in Eq.~\eqref{di}, the  average equality of odds difference (EO) in Eq.~\eqref{eo} and counterfactual fairness (CF) reported in Eq.~\eqref{cf}. For the accuracy, we calculate the Area under the ROC curve (ROC-AUC).

\begin{figure}[!htb]
\centering
\begin{minipage}{.45\textwidth}
  \centering
  \includegraphics[width=6cm]{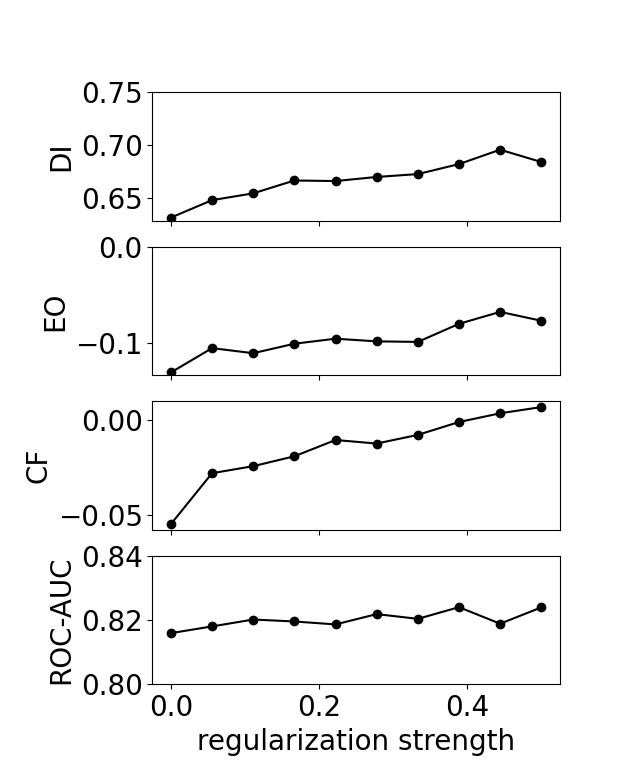}
  \captionsetup{labelformat=empty}
\end{minipage}%
\begin{minipage}{.45\textwidth}
  \centering
  \includegraphics[width=5.8cm]{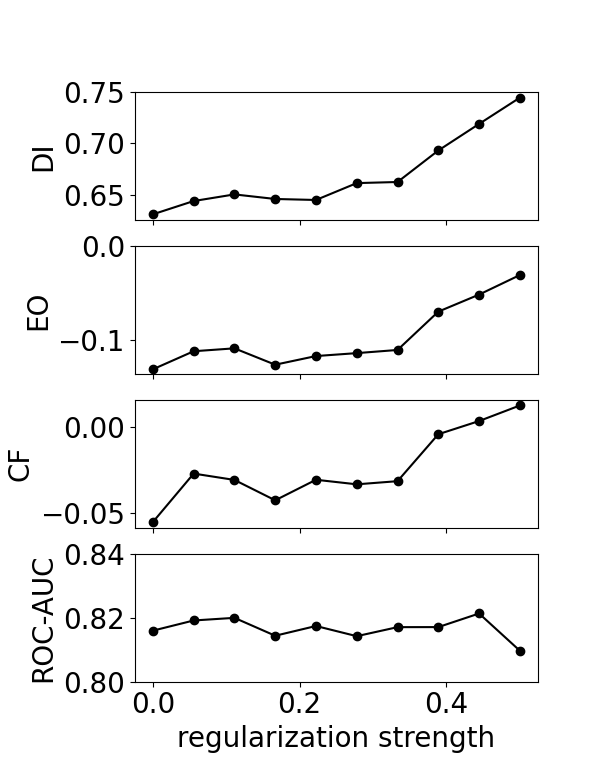}
  \captionsetup{labelformat=empty}
\end{minipage}
\caption[Fairness and accuracy metrics for Constraints to feature importance methodology]{Accuracy (ROC-AUC); and fairness measured as disparate impact (DI), average equality of odds difference (EO) and counterfactual fairness (CF) (for each measure, the higher the values the higher the fairness levels), by constraining only the \textit{race} feature (left Panel) and by constraining both \textit{race} and the correlated ones (right Panel), as a function of the regularization strength.}
\label{incr}
\end{figure}

Firstly, we evaluated the different fairness metrics in the testing set, with an increasing level of the regularization strength $\lambda$ ($10$ values from $0$ to $0.5$). In Figure~\ref{incr} we report the accuracy metric (ROC-AUC in the lower plot) and the three measures of fairness: disparate impact (DI), average equality of odds difference (EO), and counterfactual fairness (CF) as a function of the regularization strength.

In Figure~\ref{incr} we observe that, while the level of the ROC-AUC score practically remains the same, both DI, EO, and CF grow as the regularization strength augments, denoting an increased level of all the fairness measures. In particular, the CF measure reaches the value of $0$, meaning that the protected feature no longer affects the predictions. The other two measures, DI and EO, do not reach the maximum level ($1$ and $0$ respectively) because of the issue of correlated features. However, when also the correlated features are constrained through Eq.~\eqref{corr}, we note that the increase of fairness is more pronounced (right panel of Figure~\ref{incr}).\\\\
Then, as a further analysis we compared the fairness/accuracy levels obtained with the CTFI methodology\footnote{With regularization chosen to be $0.1$ and constraining also the other features, by using regularization strengths given by $\lambda \cdot \rho_{\text{gender}, i}$ (see Eq.~\eqref{corr})} to the following benchmark methodologies:
\begin{enumerate}
\item the \textit{unawareness} method \cite{grgic2016case}, avoiding to use the \textit{race} feature during the training phase;
\item a pre-processing method based on the \textit{undersampling} of the samples with protected attribute;
\item the pre-processing method called \textit{reweighing} \cite{kamiran2012data} that assigns weights to the samples in the training dataset to reduce bias.
\end{enumerate}

\begin{figure}[!htb]
\centering
  \includegraphics[width=8cm]{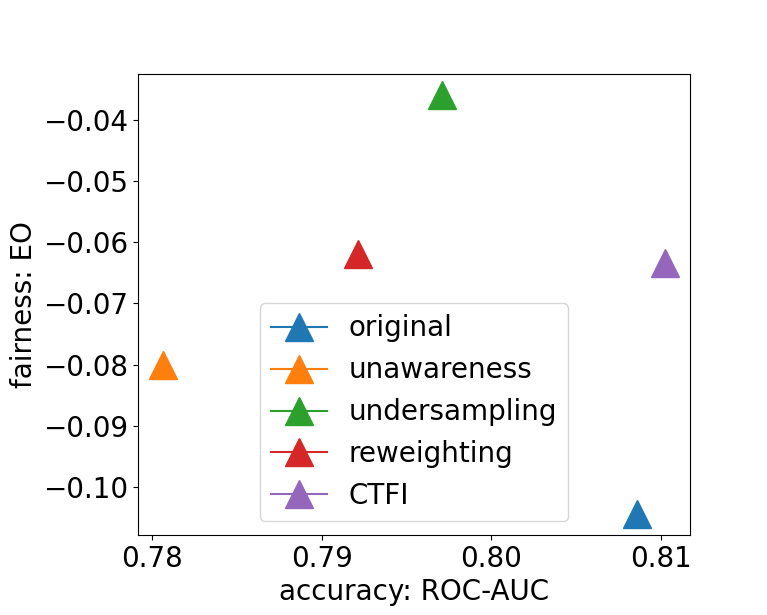}

\caption{Results of accuracy (ROC-AUC) and fairness (EO) for the different methodologies: original, unawareness, undersampling, reweighting and CTFI.}
\label{res_}
    \end{figure}

The AIF-360 library was used to apply the benchmark methodologies and the fairness metrics. All the models are coded in the Pytorch environment and available at the Github repository \url{https://github.com/nicolapicchiotti/ctfi}.\\\\
In Figure~\ref{res_} we report the results of the accuracy measure given by the ROC-AUC (x-axis) and fairness metric EO (y-axis) for the different methodologies (unawareness, undersampling, reweighting) and the CFTI. The values are reported in Table~\ref{res__}. 

\begin{table}[!htb]
\centering
\begin{tabular}{c|c|c}
    method & \multicolumn{1}{c|}{ROC-AUC} & EO \\
    \midrule
    original & 0.809 & -0.104 \\
    unawareness & 0.781 & -0.080 \\
    undersampling & 0.797 & -0.036 \\
    reweighting & 0.792 & -0.062 \\
    CTFI  & 0.810 & -0.063 \\
      \end{tabular}%
\caption{Results of the trade-off between accuracy (ROC-AUC) and fairness (EO) for the different methodologies: original, unawareness, undersampling, reweighting and CTFI.}
\label{res__}
\end{table}

We note that, with respect to the original model, the unawareness, the undersampling, and the reweighting methodologies grant a high level of fairness at the expense of accuracy. On the other side, the CTFI methodology provides higher fairness metrics with a similar level of accuracy.

\section{Conclusion and future work}
In this work we have presented a novel model agnostic framework able to inject the apriori knowledge on the relevance of the input features into a machine learning model. This "weighted" approach can contribute to bridge the gap between the fully data-driven models and the human-guided ones. The advantage of the proposed method is the flexibility:  the logic constraints are fully customizable and do not depend neither on the nature of input features (numerical, categorical etc.) nor on the architecture of the model, nor on the algorithms chosen for the computation of the feature importance, e.g. SHAP, LIME, LRP, etc.

\begin{figure}[h!]
    \begin{center}
        \includegraphics[width=45mm]{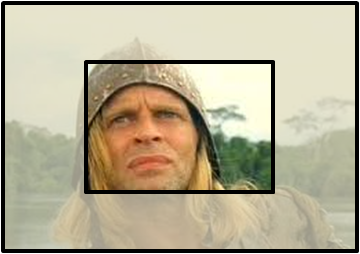}
    \end{center}
    \caption{The middle part of an image may be subject to an apriori focus.}
    \label{kin}
\end{figure}

A further application of the proposed framework is to enforce an apriori selective attention of the model on particular features, e.g. $I_i(\boldsymbol{x}) > c_i$.
This can be useful for example when the user wants to focus on some relevant words in the text, or a region of an image (see Figure \ref{kin}).

Furthermore, there could be many cases where the users want to inject prior knowledge in the form of feature importance in the model. For example, from experience, one could know that one feature should be less important than another for the business of the company, e.g. the age, the gender in a particular financial context. Also in the medical field, the a priori knowledge of the input features' importance can improve model performances where the sample sizes are limited. Another possibility is when we apriori know which feature is less reliable e.g. less stationary with respect to the others.\footnote{In linear regression a similar problem is called attenuation bias,  
where errors in the input features cause the weights going toward zero.}  
It is worth noting that the constraints can be settled for just a portion of the dataset.

As future work we are interested in providing a software solution for the integration of the proposed framework within the popular machine learning software. Another future work is to apply the logic constraints to other contexts, in terms of both dataset, e.g. images, text etc. and models (random forest, SVM etc.). Finally, the usage of other measures based on information entropy can be explored in order to take into account the problem of correlation between features.

\printbibliography
\end{document}